\documentclass{article}

\usepackage{arxiv}

\usepackage{amsmath,amssymb,amsfonts}
\usepackage{bm}

\usepackage{graphicx}
\usepackage{booktabs}
\usepackage{multirow}
\usepackage{array}
\usepackage{subcaption}
\usepackage{float}

\usepackage{algorithm}
\usepackage{algorithmic}

\usepackage{hyperref}
\usepackage{url}
\usepackage{cite}

\usepackage{listings}
\usepackage{xcolor}

\usepackage{tikz}
\usetikzlibrary{shapes.geometric, arrows.meta, positioning, calc, fit, backgrounds}
\definecolor{hotcolor}{HTML}{E07A5F}    
\definecolor{coldcolor}{HTML}{4A90D9}   
\definecolor{neutralcolor}{HTML}{E8E8E8} 

\title{A Collision-Free Hot-Tier Extension for Engram-Style Conditional Memory:\\A Controlled Study of Training Dynamics}

\author{
    Tao Lin\thanks{This work was completed in a personal capacity and does not represent the views of any employer.} \\
    \texttt{nblintao@gmail.com}
}

\date{January 23, 2026}

\begin{document}

\maketitle
\setcounter{footnote}{0}

\begin{abstract}
We investigate whether high-frequency key collisions are a primary bottleneck in Engram-style conditional memory. To isolate the effect of collisions, we introduce \textit{Engram-Nine}, a collision-free hot-tier extension that maps the most frequent n-grams through a Minimal Perfect Hash Function (MPHF) while retaining the original multi-head hashed lookup as a cold tier. Under a strictly iso-parameter setup, the collision-free design does not consistently improve validation loss.

Through route-stratified evaluation—decomposing per-token loss into hot/cold contributions—we uncover a consistent \textit{hot-to-cold advantage flip} during training: hot (high-frequency) positions initially have lower loss, but cold positions eventually surpass them. Crucially, collision-free configurations flip earlier than collision-prone baselines, suggesting that collisions act as implicit regularization. We also identify a \textit{gating mismatch}: the gate learns to favor hot positions early in training, but this preference persists even after the flip, assigning higher weights to positions with higher loss.

Our findings suggest that improving lookup precision alone does not guarantee better training outcomes. The dominant limitation may lie in gating credit assignment rather than index accuracy, and collision-induced noise may provide beneficial regularization that should not be naively eliminated.
\end{abstract}

\section{Introduction}

\subsection{Background and Motivation}

The parameter scale of large language models (LLMs) continues to grow, yet research has shown that a considerable portion of model capacity is devoted to memorizing ``local, static, and highly stereotyped'' patterns~\cite{engram2026}. The Engram module proposed by DeepSeek~\cite{engram2026} offloads these static patterns from the Transformer's attention mechanism to an external memory module via O(1) n-gram hash lookup, thereby freeing backbone capacity for more complex reasoning tasks. The appeal of this approach lies in its deterministic indexing: retrieval indices depend solely on the input token sequence, enabling index pre-computation and embedding prefetching on the host side during inference, offloading of embedding tables to host memory or NVMe storage, and prefetching of multiple candidate paths during speculative decoding.

However, the original Engram's multi-head hashing mechanism has an apparent ``weakness'': \textbf{hash collisions}. When two semantically different n-grams map to the same slot, they are forced to share the same embedding row, effectively injecting incorrect prior information into the model. Since n-gram access follows a Zipf distribution~\cite{zipf}—a small number of high-frequency keys account for the majority of queries—collisions on high-frequency keys are encountered far more often, amplifying their cumulative impact on the model. A natural intuition is: eliminating collisions for high-frequency keys should lead to lower loss.

Based on this hypothesis, we implemented \textbf{Engram-Nine}: for the top-N most frequent n-grams, we use a Minimal Perfect Hash Function (MPHF) to build a deterministic, collision-free hot tier, while long-tail cold keys continue to use the original multi-head hashing. Our goal is to answer a seemingly simple but empirically unverified question: under parameter-controlled conditions, does eliminating high-frequency collisions actually improve training outcomes?

\subsection{Main Contributions}

We report a \textbf{counterintuitive but reproducible} result: under strict iso-parameter (i.e., both methods use exactly the same number of trainable parameters) control, replacing the high-frequency n-gram index from ``high-collision hashing'' to ``collision-free MPHF (hot tier)'' does not improve validation loss. This result suggests that in Engram-style hashed memory, high-frequency collisions are not necessarily the dominant bottleneck; optimizing index precision alone is insufficient to guarantee training benefits.

To explain this phenomenon, we propose a route-stratified evaluation: on the same model checkpoint, we decompose each position's loss by hot/cold tier, combined with gate weight $\alpha$ bucketing and ablation experiments to analyze training dynamics, thereby advancing the question from ``is there benefit'' to a mechanistic diagnosis of ``why no benefit.''

The main contributions of this paper include: (1) the first iso-parameter comparison between collision-prone and collision-free variants of Engram, providing rigorous controlled experimental evidence; (2) a diagnostic methodology combining hot/cold loss decomposition with alpha bucket analysis, offering methodological reference for debugging similar systems; (3) mechanism-level discoveries including the hot/cold flip phenomenon, implicit regularization effects of collisions, and gating mismatch issues; (4) empirical evidence cautioning that improving retrieval precision does not necessarily improve training benefits—deeper mechanistic understanding is needed.

\section{Related Work}

\subsection{External Memory and Retrieval Augmentation}

Neural network external memory mechanisms have a long research history. Early Neural Turing Machines~\cite{graves2014neural} and Memory Networks~\cite{weston2015memory} explored differentiable read-write mechanisms, but their soft attention mechanisms have high computational costs at scale. Product Key Memory~\cite{lample2019large} achieves O($\sqrt{N}$) approximate retrieval through product quantization, demonstrating the value of external memory in language modeling. In recent years, Retrieval-Augmented Generation (RAG) series works~\cite{lewis2020retrieval,borgeaud2022retro} enhance generation capability by retrieving external documents. A key distinction is that RAG operates at \textbf{inference time}---dynamically retrieving from external corpora during generation---while Engram integrates memory at \textbf{training time}, learning to utilize n-gram statistics as the model trains. Additionally, RAG's retrieval granularity is at the document or passage level, whereas Engram operates at token-level n-gram precision.

The uniqueness of Engram~\cite{engram2026} lies in its \textbf{determinism}: retrieval indices are completely determined by the input token sequence, requiring no learned query encoding or soft attention computation. This makes it particularly suitable for prefetching and offload optimization during inference. Our work focuses on the collision problem within Engram, rather than comparison with other external memory architectures.

\subsection{Hashing and Perfect Hashing}

Hash collisions are an inherent problem of traditional hash tables. Cuckoo Hashing~\cite{pagh2004cuckoo} achieves O(1) worst-case query through multi-table probing, but has lower space efficiency (load factor $\sim$0.5). Minimal Perfect Hash Functions (MPHF)~\cite{fks1984} provide collision-free mapping $[0, N-1]$ for static key sets—that is, $N$ keys are mapped to exactly $N$ consecutive slots with no collisions and no wasted space, achieving the theoretical minimum storage for a bijective hash. In recent years, tools like BBHash~\cite{bbhash} and PTHash~\cite{pthash} have made MPHF construction efficient and practical (approximately 2--4 bits/key, depending on algorithm and parameters). We chose MPHF as the indexing scheme for the hot tier precisely because of its compactness and determinism.

However, a known property of MPHF is that it also returns a ``pseudo-index'' for non-member keys, thus requiring an additional membership test (such as fingerprint comparison). This extra overhead is one reason for Engram-Nine's throughput reduction.

\subsection{Regularization and Noise Injection}

Regularization techniques in machine learning—Dropout~\cite{srivastava2014dropout}, Label Smoothing~\cite{szegedy2016rethinking}, data augmentation, etc.—prevent overfitting by introducing controlled noise. Our experimental results suggest that hash collisions may play a similar role: they force semantically similar n-grams to share embeddings, equivalent to a form of implicit clustering or averaging. This observation echoes the findings of Hash Embeddings~\cite{svenstrup2017hash}, which achieves comparable performance to traditional embeddings despite hash collisions causing different tokens to share embedding vectors.

\section{Methods}

\subsection{The Original Engram's Multi-head Hash Tier}

To facilitate understanding of Engram-Nine's design motivation and technical details, this section first briefly introduces the core mechanism of the original Engram. The Engram module performs two-stage operations at each token position $t$: \textbf{retrieval} and \textbf{fusion}.

\paragraph{Retrieval Stage} The module computes deterministic indices from suffix n-grams and obtains static memory vectors through O(1) table lookup.

First is \textbf{Tokenizer Compression}: standard subword tokenizers often assign different IDs to semantically equivalent tokens (e.g., ``Apple'' and ``apple'', ``\textbackslash n'' and ``\textbackslash t''). Engram pre-computes a mapping from original token IDs to canonical IDs based on text equivalence (NFKC normalization, lowercasing, etc.) to improve semantic density.

Then is \textbf{Multi-head Hashing}: directly parameterizing the combinatorial space of all possible n-grams is infeasible—even for 2-grams alone, a 128K vocabulary would produce $128\text{K}^2 \approx 16\text{B}$ possible combinations. Therefore, Engram uses hashing to map n-grams to fixed-size embedding tables.

However, a single hash function inevitably produces \textbf{collisions}: semantically different n-grams may map to the same slot, forced to share embeddings. To mitigate this, Engram uses $K$ independent hash heads for each n-gram order $n$. The design intuition is: even if one head has a collision, other heads may still provide discriminative information, and the concatenated result of multiple heads is more expressive than a single head.

Specifically, for each order $n$ (e.g., 2, 3), take the suffix n-gram $g_{t,n} = (x'_{t-n+1}, \ldots, x'_t)$. For each head $k \in \{1, \ldots, K\}$, use a deterministic hash function $\phi_{n,k}$ (implemented as lightweight multiply-xor hash) to compute the index and perform lookup:
\begin{equation}
    z_{t,n,k} = \phi_{n,k}(g_{t,n}), \quad e_{t,n,k} = E_{n,k}[z_{t,n,k}]
\end{equation}
where $E_{n,k}$ is the corresponding embedding table (size typically chosen as a prime to improve hash distribution). Finally, concatenate vectors from all orders and all heads into the memory vector:
\begin{equation}
    e_t = \underset{n}{\text{Concat}}\left(\underset{k}{\text{Concat}}(e_{t,n,k})\right) \in \mathbb{R}^{d_{\text{mem}}}
\end{equation}

\paragraph{Fusion Stage} Engram uses context-aware gating to inject the retrieved static memory into the residual stream. Since $e_t$ is a purely local n-gram-based static prior that may contain noise due to hash collisions or polysemy, the current context is needed for dynamic modulation.

Specifically, let $h_t \in \mathbb{R}^d$ be the hidden state at the current position (having aggregated global context information through preceding attention layers), serving as the dynamic Query. The retrieved memory $e_t$ is projected to Key and Value through learnable projection matrices $W_K, W_V \in \mathbb{R}^{d \times d_{\text{mem}}}$:
\begin{equation}
    k_t = W_K \cdot e_t, \quad v_t = W_V \cdot e_t
\end{equation}
Then compute the scalar gating weight $\alpha_t \in (0, 1)$:
\begin{equation}
    \alpha_t = \sigma\left(\frac{\text{RMSNorm}(h_t)^\top \cdot \text{RMSNorm}(k_t)}{\sqrt{d}}\right)
\end{equation}
where $\sigma$ is the sigmoid function, RMSNorm is a normalization method (for gradient stability), $d$ is the hidden state dimension, and the $\sqrt{d}$ scaling factor prevents overly large dot products. The final output is:
\begin{equation}
    \text{output} = \alpha_t \cdot v_t
\end{equation}

The gating design intent is to achieve semantic alignment: the dot product between $h_t$ (Query) and $k_t$ (Key) measures the ``consistency'' between the current context and the retrieved memory. If they semantically match, the dot product is large, $\alpha_t$ approaches 1, and memory injection is stronger; if $e_t$ contradicts the current context (e.g., collision causes retrieval of irrelevant embedding), then $\alpha_t$ approaches 0, effectively suppressing noise.

\subsection{Engram-Nine's Two-Tier Retrieval Architecture}

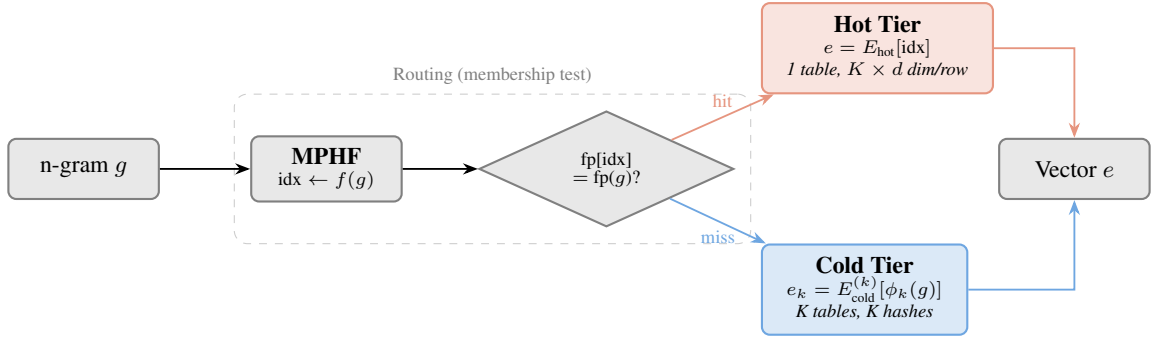
\begin{figure}[htbp]
\centering
\begin{tikzpicture}[
    box/.style={rectangle, rounded corners=3pt, minimum width=2cm, minimum height=0.8cm,
                draw=black!70, line width=0.8pt, font=\small},
    hotbox/.style={box, fill=hotcolor!20, draw=hotcolor!80},
    coldbox/.style={box, fill=coldcolor!20, draw=coldcolor!80},
    neutralbox/.style={box, fill=neutralcolor, draw=black!50},
    smallbox/.style={rectangle, rounded corners=2pt, minimum width=1.2cm, minimum height=0.5cm,
                     draw=black!50, line width=0.6pt, font=\scriptsize},
    decision/.style={diamond, aspect=2.2, minimum width=1.8cm,
                     draw=black!50, line width=0.8pt, fill=neutralcolor, font=\small},
    arrow/.style={-{Stealth[length=2mm]}, line width=0.7pt},
    hotarrow/.style={arrow, hotcolor!80},
    coldarrow/.style={arrow, coldcolor!80},
]
\node[neutralbox] (input) {n-gram $g$};

\node[neutralbox, right=1.2cm of input] (mphf) {\begin{tabular}{c}\textbf{MPHF}\\[-2pt]\scriptsize $\text{idx} \leftarrow f(g)$\end{tabular}};

\node[decision, right=1cm of mphf] (check) {\begin{tabular}{c}\scriptsize fp[idx]\\[-3pt]\scriptsize $=$ fp($g$)?\end{tabular}};

\node[hotbox, above right=0.6cm and 1.2cm of check, minimum height=1.2cm] (hot) {
    \begin{tabular}{c}\textbf{Hot Tier}\\[-2pt]\scriptsize $e = E_{\text{hot}}[\text{idx}]$\\[-2pt]\scriptsize\textit{1 table, $K \times d$ dim/row}\end{tabular}};

\node[coldbox, below right=0.6cm and 1.2cm of check, minimum height=1.2cm] (cold) {
    \begin{tabular}{c}\textbf{Cold Tier}\\[-2pt]\scriptsize $e_k = E^{(k)}_{\text{cold}}[\phi_k(g)]$\\[-2pt]\scriptsize\textit{K tables, K hashes}\end{tabular}};

\node[neutralbox, right=3.5cm of check] (output) {Vector $e$};

\draw[arrow] (input) -- (mphf);
\draw[arrow] (mphf) -- (check);
\draw[hotarrow] (check) -- node[above, font=\scriptsize] {hit} (hot);
\draw[coldarrow] (check) -- node[below, font=\scriptsize] {miss} (cold);
\draw[hotarrow] (hot) -| (output);
\draw[coldarrow] (cold) -| (output);

\begin{scope}[on background layer]
    \node[draw=gray!40, dashed, rounded corners=5pt, fit=(mphf)(check), inner sep=6pt,
          label={[font=\scriptsize, gray]above:Routing (membership test)}] {};
\end{scope}
\end{tikzpicture}
\caption{Engram-Nine two-tier retrieval architecture. Hot tier provides collision-free indexing via MPHF, cold tier retains the original multi-head hashing.}
\label{fig:architecture}
\end{figure}

The core idea of Engram-Nine is tiered processing (see Figure~\ref{fig:architecture}): use MPHF to provide collision-free indexing for high-frequency n-grams, while long-tail cold keys continue to use the original multi-head hashing. For a static set $S$ (top-N n-grams), MPHF provides a mapping $f: S \rightarrow \{0, \ldots, |S|-1\}$, collision-free for keys in $S$. But since MPHF only guarantees collision-free mapping for member keys, for non-member keys $\text{key} \notin S$, $f(\text{key})$ still returns a ``pseudo-index'', thus requiring an additional membership test.

\begin{algorithm}[htbp]
\caption{Engram-Nine Two-Tier Retrieval}
\begin{algorithmic}[1]
\REQUIRE n-gram key $g$, order $n$, heads $K$, dim $d$
\STATE $\text{idx} \leftarrow \text{mphf}[n].\text{query}(g)$
\STATE $\text{fp}_g \leftarrow \text{fingerprint}(g)$
\IF{$\text{fp\_table}[n][\text{idx}] = \text{fp}_g$}
    \STATE $e \leftarrow E_{\text{hot}}[n][\text{idx}]$ \quad // Hot: 1 table
\ELSE
    \FOR{$k = 1$ to $K$}
        \STATE $\text{idx}_k \leftarrow \phi_k(g) \mod |\text{table}|$ \quad // Cold: $K$ tables
        \STATE $e_k \leftarrow E_{\text{cold}}^{(k)}[n][\text{idx}_k]$
    \ENDFOR
    \STATE $e \leftarrow [e_1; e_2; \ldots; e_K]$ \quad // concat
\ENDIF
\RETURN $e$
\end{algorithmic}
\end{algorithm}

Engram-Nine's design goal is to verify the effect of ``eliminating high-frequency collisions'' while maintaining compatibility with the original. The key difference between the two paths lies in the indexing mechanism:

\textbf{Hot tier}: Uses MPHF to obtain a unique index idx, retrieving a complete $K \times d$ dimensional vector from a single embedding table $E_{\text{hot}} \in \mathbb{R}^{N_{\text{hot}} \times Kd}$. Since MPHF is collision-free for the hot key set, each hot key has its own dedicated row.

\textbf{Cold tier}: Follows the original Engram's multi-head hashing. For the same n-gram, use $K$ different hash functions $\phi_1, \ldots, \phi_K$ to compute indices separately, retrieving one $d$-dimensional vector from each of $K$ independent tables $E^{(k)}_{\text{cold}} \in \mathbb{R}^{M \times d}$, concatenating to obtain $K \times d$ dimensional output. Due to hash collisions, different n-grams may share the same row.

Both paths output the same dimension ($K \times d$), ensuring module interface consistency for ablation study attribution.

A key design constraint is: \textbf{the hot tier membership set is static}. Specifically, we perform a one-time frequency count on the corpus before training, determining the top-N most frequent n-grams as hot tier members, which then remain unchanged throughout training and inference. This constraint aligns with MPHF's inherent properties—MPHF can only be built for static key sets and does not support dynamic insertion or deletion. Because of this, Engram-Nine fully preserves the original Engram's core inference advantages: retrieval indices still depend only on the input token sequence, can be pre-computed and prefetched on the host side, support offloading embedding tables to host memory or NVMe storage, and are compatible with multi-path prefetching during speculative decoding. In other words, introducing the MPHF hot tier does not change Engram's deterministic indexing property—it only changes high-frequency key indexing from ``with collisions'' to ``without collisions.''

\section{Experimental Setup}

\subsection{Training Scale and Data}

Our experiments were conducted at medium scale, balancing experimental cost with conclusion reliability. Table~\ref{tab:config} summarizes the main configurations.

\begin{table}[htbp]
\centering
\caption{Experimental Configuration}
\label{tab:config}
\begin{tabular}{ll}
\toprule
\textbf{Configuration Item} & \textbf{Value} \\
\midrule
Dataset & FineWeb-Edu~\cite{fineweb} (100M tokens) \\
Tokenizer & DeepSeek-V3 (128,815 vocab) \\
Backbone & GPT-2 architecture ($\sim$185M params$^{\dagger}$, 12 layers, 768 dim) \\
Sequence length & 1024 tokens \\
Engram layers & Layer 2, 4, 6 \\
n-gram orders & [2, 3] \\
Heads per order (K) & 2 \\
Embedding dim per head & 64 \\
\bottomrule
\multicolumn{2}{l}{\footnotesize $^{\dagger}$Larger than standard GPT-2 (124M) due to expanded vocabulary (128,815 vs 50,257).}
\end{tabular}
\end{table}

\paragraph{Parameter Sharing Across Layers} The three Engram layers (Layer 2, 4, 6) share a single set of n-gram embedding tables to avoid parameter explosion. However, each layer maintains its own independent gating parameters ($W_Q$, $W_K$, $W_V$, RMSNorm), depthwise convolution, and output projection. This design keeps the model compact while allowing layer-specific modulation---as evidenced by the per-layer gating differences analyzed in Section~\ref{sec:gate-analysis}.

\paragraph{Training Hyperparameters} Training uses the nanoGPT~\cite{nanogpt} framework with AdamW optimizer ($\beta_1 = 0.9$, $\beta_2 = 0.95$, weight decay = 0.1). Learning rate is set to $6 \times 10^{-4}$ with cosine annealing schedule (warmup 100 steps, decay to $6 \times 10^{-5}$). Batch size is 4, gradient accumulation 4 steps, effective batch size 16. Total training is 5000 steps (approximately 82M tokens). Each configuration is run with 3 different random seeds to assess variance. Hardware is a single NVIDIA A100-SXM4-40GB GPU.

\paragraph{Data Split} Validation set is the last 5\% of token sequences (approximately 5M tokens), training set is the first 95\% (approximately 95M tokens). Evaluation is performed every 250 steps with 100 eval iterations per batch.

\subsection{Iso-Parameter Comparison Design}

To ensure fair comparison, Hash baseline and Engram-Nine use exactly the same total parameter count (313,567,232). The only difference between them is: \textbf{whether high-frequency n-gram indexing is collision-free}. This design allows us to isolate the effect of the single variable ``collision.''

\begin{table}[htbp]
\centering
\caption{Iso-Parameter Configuration Details}
\label{tab:iso-param}
\begin{tabular}{lp{8cm}c}
\toprule
\textbf{Config} & \textbf{Engram Embedding Parameter Allocation} & \textbf{Total Params}$^*$ \\
\midrule
Hash-500K & 2 orders $\times$ 2 heads $\times$ 500K slots $\times$ 64 dim = 128M & 313,567,232 \\
\midrule
Nine-100/400K & Hot: 2 orders $\times$ 100K slots $\times$ 128 dim = 25.6M \newline
            Cold: 2 orders $\times$ 2 heads $\times$ 400K slots $\times$ 64 dim = 102.4M \newline
            Total: 128M & 313,567,232 \\
\bottomrule
\multicolumn{3}{l}{\footnotesize $^*$Total = GPT-2 backbone ($\sim$185M) + shared Engram embeddings (128M) + per-layer gating/conv/projection ($\sim$1.6M for 3 layers).}\\
\multicolumn{3}{l}{\footnotesize \phantom{$^*$}Position embeddings excluded following common practice.}
\end{tabular}
\end{table}

\paragraph{Naming Convention} In configuration names, Hash-$X$K indicates the Hash baseline uses $X$K slots for embedding tables; Nine-$H$/$C$K indicates an Engram-Nine configuration, where $H$K is the hot tier slot count and $C$K is the cold tier slots per head. For example, Nine-100/400K means hot tier 100K slots, cold tier 400K slots/head.

\paragraph{Hot Tier Member Selection} Hot tier n-gram members are determined by pre-counting frequencies on the complete training corpus (95M tokens). For 2-grams and 3-grams, statistics are computed independently, each selecting the top-$N_{\text{hot}}$ most frequent as hot tier members. The main experiment uses $N_{\text{hot}} = 100\text{K}$; ablation experiments also test 20K and 50K. Statistics show that top-100K 2-grams cover approximately 55\% of training queries, and 3-grams cover approximately 23\%. Once selected, hot tier members remain static throughout training and inference.

\subsection{Evaluation Metrics}

Beyond standard val\_loss and throughput, we introduce this paper's core \textbf{route-stratified evaluation} metrics, to decompose the surface phenomenon of ``similar overall loss'' into finer-grained behavioral differences.

\paragraph{Route Labeling (Hot-Tier Mask)} At each evaluation, for each token position, we label whether its n-gram belongs to the top-$N_{\text{hot}}$ high-frequency set. The purpose of this label is to answer: ``If this position were in Engram-Nine, would it be routed to the hot tier or cold tier?''

It should be emphasized that the Hash baseline \textbf{does not} have native hot/cold routing during training—all n-grams take the same multi-head hash path. Therefore, the Hash baseline's hot-tier label is a \textbf{post-hoc annotation}, purely for analysis purposes: it tells us, for the same batch of validation data, which positions would be routed to the hot tier in the Nine version. By using the same top-$N_{\text{hot}}$ frequency threshold for labeling both configurations, we can fairly compare: ``Under the same hot/cold division, how do the collision-prone Hash and collision-free Nine each perform?''

\paragraph{Stratified Loss} Based on the hot-tier mask, we compute separately:
\begin{itemize}
    \item \texttt{val\_loss\_hot}: Average loss for hot-tier positions
    \item \texttt{val\_loss\_cold}: Average loss for cold-tier positions
    \item \texttt{hot\_cold\_delta}: The difference between them ($\text{val\_loss\_hot} - \text{val\_loss\_cold}$)
\end{itemize}

\paragraph{Stratified Alpha} Similar to stratified loss, we compute gating weights by the hot-tier mask:
\begin{itemize}
    \item $\alpha_{\text{hot}}$: Average gating weight for hot-tier positions
    \item $\alpha_{\text{cold}}$: Average gating weight for cold-tier positions
\end{itemize}
This is used to diagnose whether gating has formed a preference for hot/cold routing, and whether this preference aligns with actual performance.

\paragraph{Alpha Bucket Analysis} We bucket tokens by gating weight $\alpha$ into 5 intervals ($[0, 0.2), [0.2, 0.4), \ldots, [0.8, 1.0]$), analyzing average loss and hot proportion within each bucket. This diagnoses the correlation between gating weights and prediction performance—ideally, high $\alpha$ positions should correspond to low loss.

\paragraph{Per-Layer Breakdown} Since the three Engram layers (Layer 2, 4, 6) each have independent gating parameters, we also record $\alpha_{\text{hot}}$ and $\alpha_{\text{cold}}$ distributions per layer, for locating potential layer-level anomalies.

\section{Experimental Results}

\subsection{Main Finding: Eliminating Collisions Yields No Significant Benefit}

We first report the main results of the iso-parameter controlled experiment. Table~\ref{tab:main-results} summarizes all configurations from main and ablation experiments, each with 1--3 random seeds.

\begin{table}[htbp]
\centering
\caption{Summary of All Experimental Configuration Results}
\label{tab:main-results}
\begin{tabular}{lccccc}
\toprule
\textbf{Config} & \textbf{val\_loss} & \textbf{std} & \textbf{hot\_cold\_delta} & \textbf{Throughput (tok/s)} & \textbf{Params} \\
\midrule
\multicolumn{6}{l}{\textit{Main Controlled Experiment}} \\
Hash-500K & 4.4809 & 0.0082 & +0.07 & $\sim$1910 & 313,567,232 \\
Nine-100/400K & \textbf{4.4799} & 0.0123 & +0.10 & $\sim$1693 & 313,567,232 \\
\midrule
\multicolumn{6}{l}{\textit{Ablation Experiments}} \\
Hash-300K & 4.4825 & -- & +0.08 & $\sim$1917 & 313,567,232 \\
Hash-800K & 4.4961 & -- & +0.08 & $\sim$1917 & 313,567,232 \\
Nine-20/480K & 4.4872 & -- & +0.17 & $\sim$1670 & 313,567,232 \\
Nine-50/450K & 4.4942 & -- & +0.12 & $\sim$1670 & 313,567,232 \\
\bottomrule
\end{tabular}
\end{table}

The core finding is: under strict iso-parameter conditions, Engram-Nine's collision-free hot tier did not bring significant val\_loss improvement. Although Nine-100/400K achieved the lowest validation loss among all configurations (4.4799), its advantage over Hash-500K is only 0.001, far smaller than the measurement standard deviation (0.008--0.012), and is not statistically significant. Ablation experiments further confirm that regardless of whether $N_{\text{hot}}$ is 20K, 50K, or 100K, Nine configurations did not significantly outperform corresponding Hash baselines. More notably, the high-collision Hash-300K (4.4825) performed comparably to the collision-free Nine-100/400K, directly challenging the intuitive assumption that ``eliminating collisions must be beneficial.''

However, the hot\_cold\_delta column in the table reveals a phenomenon worth investigating: all configurations show cold outperforming hot at the end of training (delta $>$ 0), and Nine configurations' delta is systematically larger than Hash configurations. This prompts us to go beyond overall loss into finer-grained analysis.

\paragraph{Throughput} As shown in the table, Nine configurations bring approximately 11--12\% throughput reduction, mainly from MPHF query and fingerprint comparison overhead. This overhead can be reduced through batched queries, SIMD/GPU acceleration, etc., but since this paper focuses on training effectiveness rather than system optimization, we did not pursue this direction further.

\subsection{Core Finding: Hot/Cold Flip Phenomenon}

Through route-stratified evaluation, we discovered a training dynamics phenomenon consistent across configurations: the \textbf{Hot/Cold Flip}.

\begin{figure}[htbp]
\centering
\includegraphics[width=0.95\textwidth]{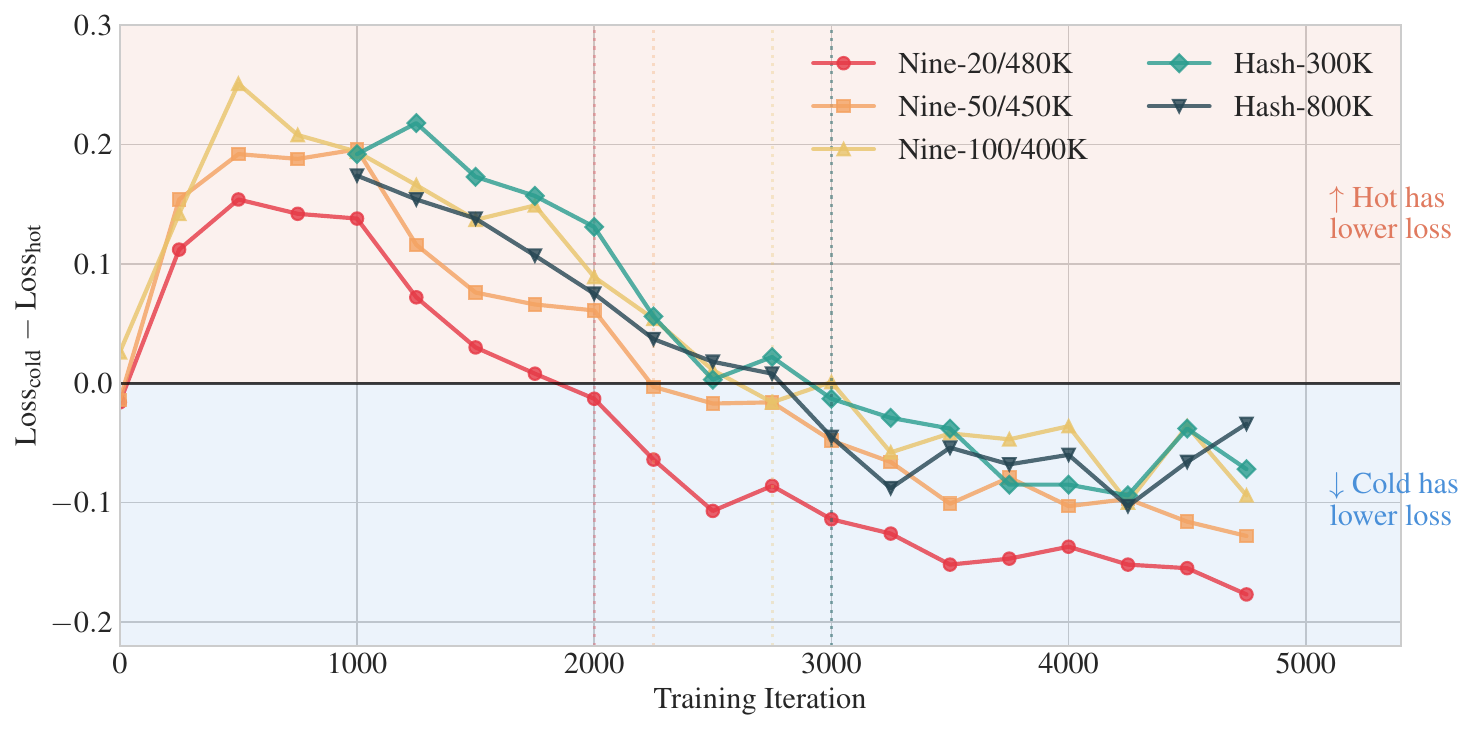}
\caption{Training dynamics of the Hot/Cold flip phenomenon. All configurations transition from early hot advantage (delta $>$ 0) to late cold advantage (delta $<$ 0). Nine configurations flip earlier than Hash configurations. (Hash-500K only preserved the final checkpoint, so it is not included in flip timing analysis.)}
\label{fig:training-dynamics}
\end{figure}

Figure~\ref{fig:training-dynamics} shows the evolution of hot\_cold\_delta ($\text{val\_loss\_cold} - \text{val\_loss\_hot}$) over training steps. In early training (iter 1000--2000), hot positions' loss is notably lower than cold positions (delta $\approx +0.15 \sim +0.22$), consistent with the intuition that ``hot keys are easier to predict''—high-frequency n-grams have more concentrated successor distributions, making them easier for the model to learn. However, by late training (iter 4000+), cold positions overtake, with the gap widening to $-0.07 \sim -0.17$.

Table~\ref{tab:flip-details} provides detailed comparison of flip timing and magnitude across configurations, revealing three key patterns.

\begin{table}[htbp]
\centering
\caption{Detailed Comparison of Flip Timing and Magnitude}
\label{tab:flip-details}
\begin{tabular}{lcccc}
\toprule
\textbf{Config} & \textbf{Collision Status} & \textbf{Flip Time} & \textbf{Max Hot Advantage Pre-Flip} & \textbf{Max Cold Advantage Post-Flip} \\
\midrule
Nine-20/480K & Collision-free & iter 2000 & $-0.154$ (iter 500) & $+0.177$ (iter 4750) \\
Nine-50/450K & Collision-free & iter 2250 & $-0.196$ (iter 1000) & $+0.128$ (iter 4750) \\
Nine-100/400K & Collision-free & iter 2750 & $-0.251$ (iter 500) & $+0.100$ (iter 4250) \\
Hash-300K & High collision & iter 3000 & $-0.218$ (iter 1250) & $+0.094$ (iter 4250) \\
Hash-800K & Low collision & iter 3000 & $-0.174$ (iter 1000) & $+0.103$ (iter 4250) \\
\bottomrule
\end{tabular}
\end{table}

\paragraph{Pattern 1: Universality of the Flip} All five configurations—whether Nine or Hash, regardless of parameter allocation—exhibited the flip from hot advantage to cold advantage. This is not an artifact of a specific configuration, but a stable training dynamic that reproduces across all configurations we tested.

\paragraph{Pattern 2: Collisions Delay the Flip} Collision-free Nine configurations flip at iter 2000--2750, while collision-prone Hash configurations uniformly flip at iter 3000, approximately 250--1000 steps later. Notably, across the two table\_size endpoints we tested (300K and 800K), although the latter's collision rate is expected to be lower (slots increased by approximately 2.7$\times$), the flip timing is identical—both at iter 3000. This observation is consistent with a ``threshold hypothesis'': the regularization effect of collisions may have some threshold, where exceeding the threshold is sufficient to delay the flip. However, since we only tested two endpoints, we cannot rule out continuous dependence of flip timing on table\_size, requiring denser table\_size sweeps for verification.

\paragraph{Pattern 3: Collision-Free Leads to More Severe Divergence} After the flip, Nine configurations' cold advantage ($+0.10 \sim +0.17$) is systematically larger than Hash configurations ($+0.07 \sim +0.08$). This means eliminating collisions not only did not help the hot tier, but actually widened the performance gap between hot/cold.

\paragraph{Pattern 4: Gating Preference Opposes Performance} Despite cold loss being lower in late training, gating $\alpha$ consistently favors hot ($\alpha_{\text{hot}} > \alpha_{\text{cold}}$ stably reproduces across seeds). In other words, the gate learned to distinguish hot/cold, but its preference direction is opposite to actual prediction performance. This key phenomenon motivates the detailed gating preference analysis in Section~\ref{sec:gate-analysis}.

\subsection{\texorpdfstring{$N_{\text{hot}}$}{N\_hot} Ablation: Refuting the Sparsity Hypothesis}

$N_{\text{hot}}$ is the slot count for the hot tier, determining how many high-frequency n-grams can obtain collision-free dedicated embeddings. A natural hypothesis is: smaller $N_{\text{hot}}$ means more training samples per hot embedding (since the hot tier only contains the most frequent n-grams, each embedding's average training samples are higher), so they should learn better and flip later. However, experimental results are exactly opposite (see Table~\ref{tab:nhot-ablation}).

\begin{table}[htbp]
\centering
\caption{$N_{\text{hot}}$ Ablation Experiment}
\label{tab:nhot-ablation}
\begin{tabular}{llcccc}
\toprule
\textbf{Config} & \textbf{$N_{\text{hot}}$} & \textbf{Hot Coverage} & \textbf{Flip Time} & \textbf{Sparsity Hypothesis Expectation} & \textbf{Actual Observation} \\
\midrule
Nine-20/480K & 20K & 41\% & iter 2000 & Latest (densest samples) & \textbf{Earliest} \\
Nine-50/450K & 50K & 51\% & iter 2250 & Medium & Medium \\
Nine-100/400K & 100K & 59\% & iter 2750 & Earliest (sparsest samples) & Latest \\
\bottomrule
\end{tabular}
\end{table}

The results directly contradict the sparsity hypothesis: the configuration with the densest samples per embedding (Nine-20K) flips \emph{earliest} rather than latest. This rules out ``insufficient training samples for hot embeddings'' as the root cause.

What alternative mechanisms might explain the observed pattern? We consider two possibilities:

\paragraph{Hypothesis A: Cold-tier collision rate.} Under iso-parameter design, smaller $N_{\text{hot}}$ yields a larger cold table with lower collision rates. Since collisions provide implicit regularization that delays the flip (Section~\ref{sec:collision-reg}), a less-collided cold tier might ``learn well'' earlier and overtake sooner. However, this explanation faces a difficulty: pure Hash configurations---effectively 100\% cold tier---flip at iter 3000, \emph{later} than all Nine configurations. This suggests the flip timing may be more strongly influenced by the presence of a collision-free hot tier than by the cold table's collision rate alone, though disentangling these factors would require additional controlled experiments (e.g., artificially injecting collisions into the hot tier).

\paragraph{Hypothesis B: Hot-tier coverage.} Smaller $N_{\text{hot}}$ implies lower hot-tier coverage (41\% vs.\ 59\%), meaning the hot tier contributes less to overall loss. Consequently, even a modest degradation in hot-tier performance produces a more pronounced flip signal in the aggregate metrics.

Both hypotheses attribute the flip timing to architectural factors (collision structure, coverage ratio) rather than to per-embedding sample density, consistent with our rejection of the sparsity hypothesis.

\subsection{Gating Preference Analysis}
\label{sec:gate-analysis}

Route-stratified evaluation reveals a core issue with the gating mechanism. We analyze from two angles: (1) temporal evolution of $\alpha$ by hot/cold split; (2) loss statistics bucketed by $\alpha$ value.

\subsubsection{Temporal Analysis: Evolution of \texorpdfstring{$\alpha$}{α} During Training}

\begin{figure}[htbp]
\centering
\includegraphics[width=0.85\textwidth]{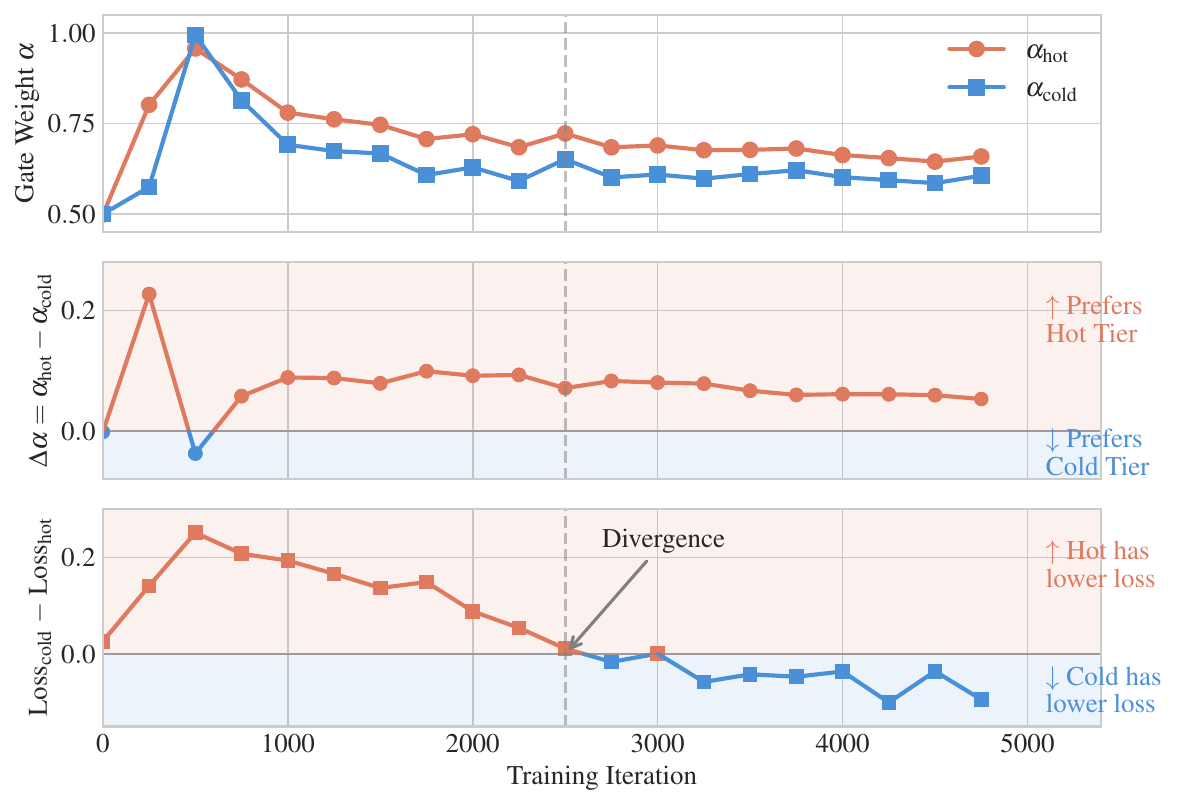}
\caption{Gate weight evolution during training (Nine-100/400K, seed0). \textbf{Top}: Raw $\alpha$ values show overall decline from $\sim$1.0 to $\sim$0.65. \textbf{Middle}: Gate preference $\Delta\alpha$ stays positive throughout, indicating persistent hot preference. \textbf{Bottom}: After iter 2500, cold achieves lower loss (negative region), but gate preference does not reverse—the preference ``crystallized'' in early training.}
\label{fig:alpha-evolution}
\end{figure}

Figure~\ref{fig:alpha-evolution} shows the evolution of raw gate weights ($\alpha_{\text{hot}}$, $\alpha_{\text{cold}}$), gate preference ($\Delta\alpha = \alpha_{\text{hot}} - \alpha_{\text{cold}}$), and actual performance difference (Loss$_{\text{cold}}$ $-$ Loss$_{\text{hot}}$) over training iterations for Nine-100/400K.

This visualization reveals several key patterns:

\paragraph{Pattern 1: Overall $\alpha$ Decline} Both hot and cold $\alpha$ decline from early $\sim$1.0 to late 0.6--0.7. This indicates the model overall becomes more ``cautious,'' with reduced trust in memory injection.

\paragraph{Pattern 2: Gate Consistently Prefers Hot} Despite the overall decline, $\alpha_{\text{hot}} > \alpha_{\text{cold}}$ always holds from iter 1000 to iter 4750, indicating the gate learned to distinguish hot/cold and formed a ``hot keys are more trustworthy'' preference. This pattern stably reproduces across all seeds.

\paragraph{Pattern 3: $\Delta\alpha$ Gradually Shrinking} While the gate maintains hot preference throughout training, the preference gap $\Delta\alpha = \alpha_{\text{hot}} - \alpha_{\text{cold}}$ is gradually shrinking. This trend is consistent across all three seeds and shows no sign of plateauing by iter 4750. This suggests that given sufficient training time, the gate might eventually correct its hot bias—but the current training duration is insufficient to observe this potential correction.

\paragraph{Pattern 4: Preference and Performance Diverge} In early training (iter 1000--2000), the gate's preference aligns with actual performance: hot loss is indeed lower, gate gives hot higher $\alpha$. However, from iter 2500 onward, cold loss becomes lower than hot loss, but the gate's preference does not reverse. In other words, the gate's preference ``crystallized'' once formed in early training, creating a mismatch between gating behavior and actual prediction performance.

\subsubsection{Snapshot Analysis: \texorpdfstring{$\alpha$}{α} Bucketing}

To further verify the mismatch between gating preference and performance, we bucket by $\alpha$ value and compute corresponding loss statistics (Table~\ref{tab:alpha-bucket}).

\begin{table}[htbp]
\centering
\caption{Alpha Bucket Analysis (Hash-500K checkpoint)}
\label{tab:alpha-bucket}
\begin{tabular}{lcccc}
\toprule
\textbf{$\alpha$ Range} & \textbf{Avg Loss} & \textbf{Hot Proportion} & \textbf{Sample Count} & \textbf{Note} \\
\midrule
0.0--0.2 & 3.94 & 27.3\% & 93,483 & \\
0.2--0.4 & \textbf{3.90} & 34.1\% & 148,173 & Lowest loss \\
0.4--0.6 & 3.94 & 47.7\% & 131,735 & \\
0.6--0.8 & 4.63 & 67.2\% & 204,112 & \\
0.8--1.0 & \textbf{5.28} & 76.2\% & 241,697 & Highest loss \\
\bottomrule
\end{tabular}
\end{table}

The data clearly confirms the temporal analysis conclusion: the model assigns higher $\alpha$ to positions with higher loss. Low $\alpha$ (0--0.4) positions have loss around 3.9, while high $\alpha$ (0.8--1.0) positions have loss as high as 5.1--5.3—completely opposite to the gating design intent. More critically, approximately 70\% of the high $\alpha$ bucket are hot positions (Hash 76\%, Nine 69\%), corroborating the finding that ``$\alpha_{\text{hot}} > \alpha_{\text{cold}}$'': gate prefers hot, but hot's loss is actually higher in late training.

\begin{figure}[htbp]
\centering
\includegraphics[width=1.0\textwidth]{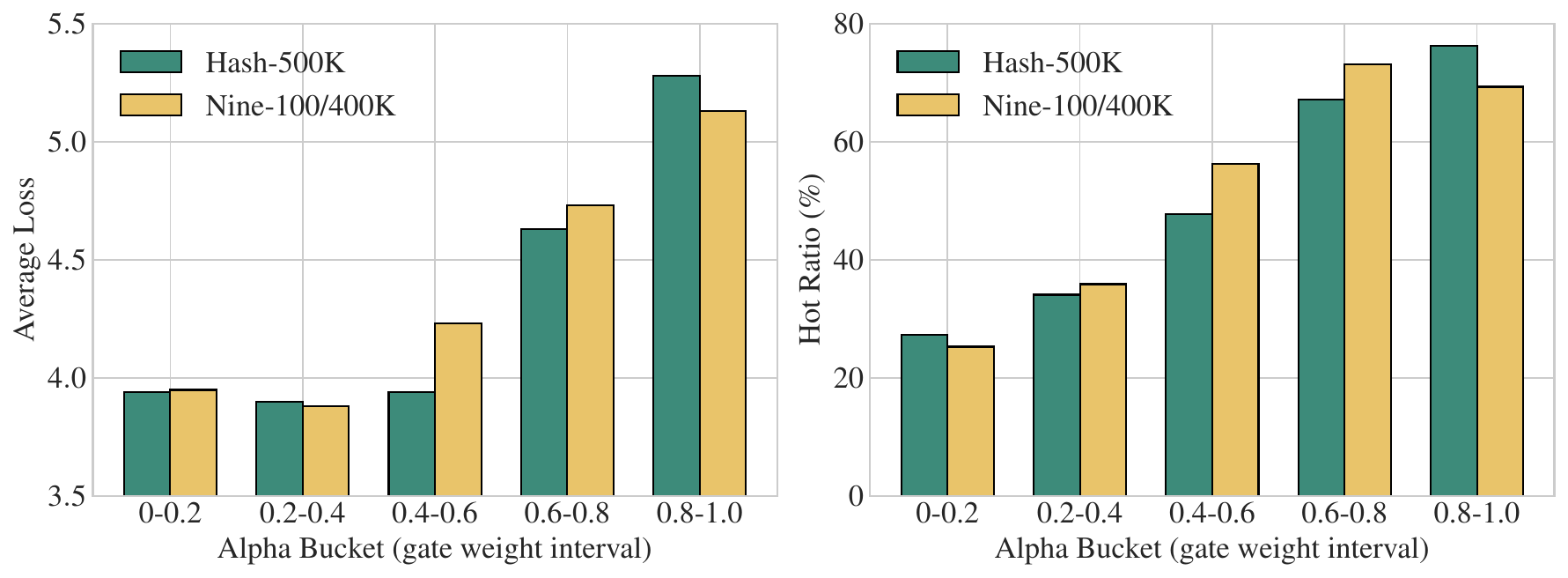}
\caption{Gate mismatch phenomenon. (a) High $\alpha$ bucket (0.8--1.0) has the highest average loss (5.1--5.3), low $\alpha$ bucket (0.2--0.4) has the lowest loss ($\approx$3.9). (b) Approximately 70\% of the high $\alpha$ bucket are hot positions.}
\label{fig:gate-mismatch}
\end{figure}

\subsubsection{Synthesis: Preference Fixation Hypothesis}

Synthesizing temporal and snapshot analyses, we propose an explanation: \textbf{gating preference forms and fixates in early training}. The specific mechanism is as follows:

\begin{enumerate}
    \item \textbf{Early stage} (iter 0--2000): Hot positions' loss is indeed lower (high-frequency n-grams are easier to predict), gate learns to ``prefer hot'' through gradients.
    \item \textbf{Middle stage} (iter 2000--3000): Cold tier gains implicit regularization through collisions, generalization ability gradually catches up with and surpasses hot tier.
    \item \textbf{Late stage} (iter 3000+): Cold loss is already lower than hot loss, but the gate's preference parameters have already ``fixated,'' failing to adjust in time.
\end{enumerate}

This mismatch appears in both Hash and Nine, indicating the problem is not in collisions themselves, but in the gating mechanism's credit assignment. Possible improvement directions include: (1) letting the gate see more information (collision degree, n-gram frequency, membership confidence); (2) using more complex gating mechanisms to enhance recalibration ability; (3) introducing some ``forgetting'' mechanism to let the gate adapt to changes in training dynamics.

\subsection{Layer-Level Anomaly: Layer 6's \texorpdfstring{$\alpha$}{α} Reversal}

\begin{figure}[htbp]
\centering
\includegraphics[width=0.85\textwidth]{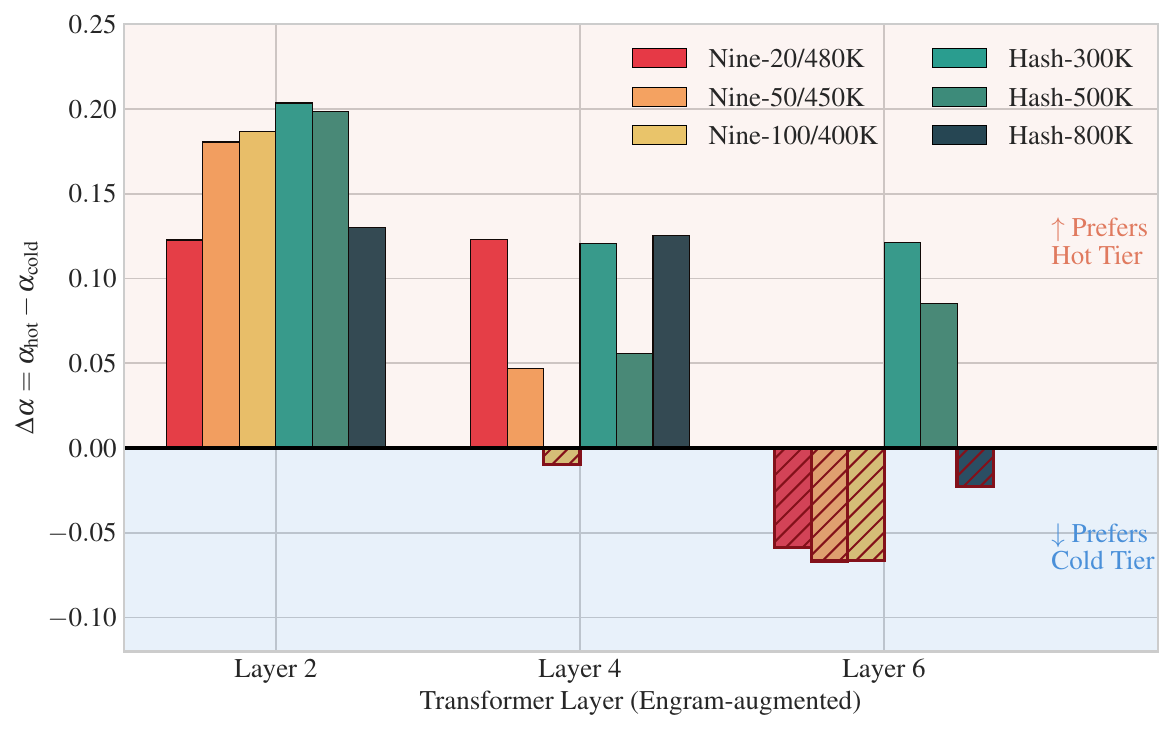}
\caption{Gating weight difference $\Delta\alpha = \alpha_{\text{hot}} - \alpha_{\text{cold}}$ across layers. Upper region (positive) indicates gate prefers hot tier, lower region (negative) indicates preference for cold tier. Hatched fill marks anomalous cold preference.}
\label{fig:layer-alpha}
\end{figure}

When analyzing gating weights per layer, we discovered a significant pattern (see Figure~\ref{fig:layer-alpha}): the gating weight difference $\Delta\alpha = \alpha_{\text{hot}} - \alpha_{\text{cold}}$ exhibits markedly different behavior across layers.

Figure~\ref{fig:layer-alpha} reveals a clear layer-level pattern:

\begin{itemize}
    \item \textbf{Layer 2 (shallow)}: All configurations' gates correctly prefer the hot tier ($\Delta\alpha > 0$).
    \item \textbf{Layer 4 (middle)}: Most configurations still prefer hot tier, but Nine-100/400K begins showing slight cold preference.
    \item \textbf{Layer 6 (deep)}: \textbf{All Nine configurations}' gates anomalously prefer the cold tier ($\Delta\alpha \approx -0.06$), Hash-800K also shows slight cold preference, while Hash-300K/500K still normally prefer hot tier.
\end{itemize}

This finding has two implications: (1) $\alpha$ reversal is strongly correlated with \textbf{collision-free design} (Nine), not directly with parameter count or hot tier size; (2) reversal is more severe in \textbf{deeper layers}, suggesting some mismatch between Engram's n-gram memory and deep-layer hidden states.

A possible mechanistic explanation is: shallow hidden states are more biased toward local/surface features (n-gram patterns themselves), naturally aligning with Engram's n-gram embeddings; deep hidden states are more biased toward global/abstract features (syntax, semantics), where the ``locality'' of n-gram memory may actually interfere with representation learning. In Hash configurations, collision-induced noise may ``blur'' this mismatch, making it harder for the gate to form incorrect preferences.

This provides a concrete entry point for future research: consider \textbf{using Engram only in shallow layers}, or adopting different gating strategies for different layers.

\section{Discussion}

\subsection{\texorpdfstring{Precision $\neq$ Better}{Precision ≠ Better}: A More General Principle}

Our experiments suggest: at this scale, eliminating collisions does not equal better training benefits. Under iso-parameter conditions, Engram-Nine's collision-free hot tier did not bring the expected loss improvement, but instead led to earlier and more severe hot/cold flipping. This result challenges the intuitive assumption that ``collisions are the main bottleneck.''

This finding may have significance beyond Engram. In machine learning system design, ``improving precision'' is often assumed to be the right direction—more precise retrieval, more accurate indexing, finer-grained representations. However, our results suggest: in certain regimes, precision improvements may be offset or even reversed by other factors (such as loss of regularization effects, changes in optimization dynamics). This observation may have loose analogies with other memory/retrieval systems that introduce ``approximation'' or ``sharing,'' such as Product Key Memory~\cite{lample2019large}'s product quantization, Hash Embeddings~\cite{svenstrup2017hash}, and quantization errors in approximate nearest neighbor retrieval—they may all serve as capacity constraints or implicit regularization to some extent. However, these methods introduce approximation through different mechanisms, and this paper's conclusions should not be directly extrapolated; specific effects need individual verification.

The practical implication for system design is: before pursuing ``more precise'' retrieval, one needs to first verify whether precision is truly the bottleneck. Adding system complexity (such as introducing MPHF) not only brings engineering overhead but may also destroy beneficial properties implicit in the original design.

\subsection{Collisions as Implicit Regularization: Mechanisms and Boundaries}
\label{sec:collision-reg}

\paragraph{Hypothesized mechanism.}
Collisions may provide an implicit regularization effect through two complementary pathways. First, collisions cause high-frequency n-gram embeddings to be ``averaged''—semantically similar but not identical n-grams share the same vector, equivalent to soft clustering. This reduces the model's risk of overfitting to specific n-grams. Conversely, the collision-free hot tier lets each embedding serve only a single n-gram, making overfitting easier. Second, from an information-theoretic perspective, collisions limit the effective capacity of the embedding table: even if the table has $M$ slots, due to collisions, the actual number of distinguishable n-grams is less than $M$. This forces the model to learn more generalizable representations rather than memorizing specific n-grams' idiosyncratic patterns—similar to the regularization principle of bottleneck structures (such as VAE's latent space).

\paragraph{Supporting observations.}
Hash configurations' flip is approximately 250--1000 iterations later than Nine, and the cold advantage after flipping is smaller. This pattern is consistent with classic regularization techniques (such as dropout, label smoothing) that delay overfitting onset.

\paragraph{Limitations and boundary conditions.}
The above observations are correlational evidence; this paper has not directly verified the causal relationship between collisions and regularization. Moreover, collisions as regularization likely have boundary conditions:
\begin{itemize}
    \item \textbf{Data scale}: When training data is sufficient, overfitting risk decreases, and the regularization value of collisions may diminish. Our 100M token experiment is in a ``relatively data-scarce'' regime, which may amplify the importance of regularization.
    \item \textbf{Collision degree}: At the two table\_size endpoints we tested (300K and 800K), flip timing is identical (iter 3000), suggesting the regularization effect of collisions may have a threshold, beyond which the specific degree has little impact. Denser sweeps are needed for verification.
    \item \textbf{N-gram semantic distribution}: The ``soft clustering'' effect depends on which n-grams collide. If colliding n-grams are semantically similar, averaging is beneficial; if semantically opposite, it introduces noise. Current random hashing cannot control this.
\end{itemize}

\subsection{Gating Credit Assignment: Why Preferences Fixate}

The detailed analysis in Section~\ref{sec:gate-analysis} reveals that the gating credit assignment problem may be the core bottleneck of the current architecture. The gating design intent is: assign higher weight to positions that ``predict accurately,'' suppress memory injection for positions that ``predict difficulty.'' However, we observe systematic mismatch: $\alpha_{\text{hot}} > \alpha_{\text{cold}}$ persists from early to late training, but the relative loss of hot/cold flips around iter 3000.

The ``preference fixation hypothesis'' proposed there can be further explained from an optimization dynamics perspective. Gating weight $\alpha$ is determined by the dot product of hidden state $h_t$ and key $k_t$, where $k_t = W_K \cdot e_t$ depends on memory embedding $e_t$. In early training, hot embeddings converge quickly due to high-frequency samples, forming stable $k_t$ representations; $W_K$ correspondingly learns to ``prefer'' these stable keys. Once the parameter space of $W_K$ is occupied by this preference, subsequent gradient updates have difficulty reversing it—because cold tier embeddings continue to change due to collisions, making it hard for $W_K$ to form stable preference for them.

Based on the above analysis, we speculate that preference fixation is not a ``bug'' of the gating mechanism, but may be a natural result of its interaction with collision dynamics—hot tier embedding stability causes $W_K$ to form early preferences, while cold tier's continuous drift makes this preference hard to reverse. This hypothesis provides two types of testable predictions:

\begin{itemize}
    \item \textbf{Stability difference}: If ``drift vs stability'' is key, then hot positions' $k_t = W_K \cdot e_t$ similarity between adjacent checkpoints should be higher than cold's; and $\alpha$'s correlation with key stability should be stronger than with loss.
    \item \textbf{$W_K$ intervention}: If ``$W_K$ early occupation'' is key, then decoupling hot/cold $W_K$ (using independent $W_K^{\text{hot}}$, $W_K^{\text{cold}}$) or applying periodic reset/EMA decay to $W_K$ should weaken mismatch or delay flipping.
\end{itemize}

\subsection{Dialogue with the Original Engram Design}

DeepSeek chose multi-head hashing rather than MPHF or other collision-free schemes in the original Engram design. The original paper explicitly acknowledges the collision problem and handles it through two mechanisms: (1) using multi-head hashing to ``mitigate collisions''; (2) using context-aware gating to suppress collision-induced noise—the paper states that ``if the retrieved memory contradicts the current context, the gate tends toward zero, effectively suppressing the noise.'' However, the original paper positions collisions as \textbf{noise to be suppressed}, not as potentially beneficial properties. Reasons for choosing multi-head hashing over collision-free schemes may include: (1) engineering simplicity—no need to pre-build static key sets; (2) dynamism—MPHF only supports static sets, while multi-head hash naturally supports arbitrary inputs.

Our experimental results provide a rationality explanation for this design choice: multi-head hashing's collisions are not only ``tolerable'' but may be ``beneficial.'' The ``noise'' that the original paper tried to suppress through gating may actually play an implicit regularization role. We cannot know whether DeepSeek was already aware of this—they may have discovered the regularization effect of collisions through internal experiments, or it may just be a serendipitous design byproduct. Regardless, this explanation needs validation at larger scales—whether the regularization effect of collisions remains significant at DeepSeek's actual training scale (hundreds of billions of tokens) is an open question.

\subsection{Recommendations for Practitioners}

Based on this paper's findings, we offer the following recommendations for practitioners who wish to use Engram or similar hashed memory systems:

\paragraph{Rationality of Retaining Collisions} At the scale tested in this paper (100M tokens, 185M backbone parameters), retaining collisions is the more robust choice. Specifically: (1) when training data is relatively limited (e.g., $<$1B tokens), the regularization effect of collisions helps prevent overfitting; (2) when model scale is smaller, overfitting risk is higher, and the implicit constraint from collisions is more valuable; (3) retaining collisions avoids introducing additional components like MPHF, making engineering simpler; (4) no membership test needed, lower inference latency. Whether eliminating collisions has value at larger scales or in specific domains requires future research validation (see Section~\ref{sec:future-work}).

\paragraph{Diagnostic Methods} Regardless of approach, we recommend using this paper's proposed route-stratified evaluation as a diagnostic tool: monitor hot/cold delta evolution, $\alpha$ bucket analysis, and per-layer gating weight breakdown. These metrics can help detect flip phenomena and gating mismatch issues early.

\section{Limitations and Future Work}

\subsection{Limitations}

\textbf{This work is a rapid proof-of-concept validation of a research idea, not a definitive conclusion.} The limitations listed below mean that actual performance at production scale may differ significantly from what we observed. The primary contribution of this paper lies in proposing the Engram-Nine architecture, the route-stratified evaluation methodology, and the analytical framework for understanding hot/cold dynamics and gating credit assignment—not in the specific numerical outcomes. With sufficient resources and time for continued research\footnote{Readers interested in funding or collaborating on follow-up research are welcome to contact the author.}, larger-scale experiments could reveal substantially different results; the collision-as-regularization effect we observed might diminish or even reverse when overfitting risk is inherently lower at scale.

Specifically, this paper's experiments have the following limitations:

\paragraph{Scale Differences} This paper's experiments were conducted at 185M backbone parameters, 100M tokens scale, while DeepSeek's original Engram was trained on tens of billions of parameters, hundreds of billions of tokens. Additionally, our Engram embedding capacity (128M) represents $\sim$69\% of the backbone (185M), a ratio likely much higher than in practical large-scale deployments where Engram serves as a lightweight auxiliary. This may amplify overfitting tendencies and the regularization benefit of collisions. At larger scales with more balanced Engram-to-backbone ratios, the collision effects we observed may differ significantly.

\paragraph{Data and Domain} We only use the FineWeb-Edu dataset; other domains (code, mathematics, multilingual) may show different behavior, as n-gram frequency distributions and semantic characteristics may differ significantly.

\paragraph{Architecture Variants} We only test the GPT-2 dense architecture. DeepSeek's original Engram uses an MoE architecture, where sparse activation may form some synergy or competition with Engram's n-gram routing.

\subsection{Future Work}
\label{sec:future-work}

Based on this paper's findings, we propose the following research directions:

\paragraph{Verifying the Regularization Mechanism of Collisions} This paper observes collisions delaying flipping and improving generalization, but has not directly verified the causal relationship. Future experiments to test the ``collision=regularization'' hypothesis include: (1) applying explicit regularization to collision-free hot embeddings (dropout, weight decay, Gaussian noise), observing whether it can reproduce Hash configurations' delayed flip effect; (2) designing ``controlled collision'' mechanisms, intentionally letting semantically similar n-grams share embeddings, verifying the ``soft clustering'' hypothesis; (3) tracking individual n-gram embedding trajectories during training, directly observing how collisions affect representation learning.

\paragraph{Distinguishing Index Effects from Architecture Effects} Current experiments cannot fully distinguish the contributions of ``collision-free indexing'' versus ``hot/cold tiered architecture.'' A key ablation is the membership-only experiment: retain hot/cold routing split, but hot tier still uses multi-head hashing (only use MPHF for membership test to decide routing, not for indexing). If this configuration still shows early flipping, the problem is in the tiered architecture itself; if flipping is delayed, the problem is indeed in collision-free. This experiment can cleanly answer: does Engram-Nine's ``no benefit'' come from MPHF's collision-free property, or from training dynamics changes brought by introducing hot/cold separation architecture.

\paragraph{Improving Gating Credit Assignment} Gating preference fixation is one of the core issues revealed in this paper. Improvement directions include: (1) enhancing gating input—letting the gate see more information, such as explicit hot-tier indicators, n-gram frequency, collision degree estimates, or membership confidence; (2) decoupling hot/cold gating parameters—using independent $W_K$ for both paths, avoiding early preference fixation affecting late adaptation; (3) introducing dynamic calibration mechanisms—such as periodic reset of gating parameters, or using EMA decay to let early preferences gradually ``forget''; (4) designing auxiliary losses—explicitly penalizing mismatch between gating weights and actual prediction performance.

\paragraph{Exploring Layer-Level Strategies for Engram} Layer 6's $\alpha$ reversal phenomenon suggests Engram may have mismatch with Transformer deep layers. Future exploration could include: (1) ablation experiments—removing Engram from Layer 6, using only Layers 2 and 4, observing whether it improves overall performance; (2) layer-level differentiation—using different hot/cold ratios or different gating strategies for different layers; (3) analyzing hidden state characteristics—studying how shallow (local-biased) versus deep (global/abstract-biased) semantic differences interact with n-gram memory.

\paragraph{Scaling to Larger Scales and More Domains} This paper's 100M token experiment is in a ``relatively data-scarce'' regime, which may amplify the regularization effect of collisions. A key question is: at 10B+ tokens, 1B+ parameter scale, does the regularization value of collisions remain significant? Relatedly, our near 1:1 Engram-to-backbone parameter ratio is likely unrealistic for production systems; exploring smaller ratios (e.g., 1:4, 1:8) may reveal different collision dynamics or even benefits from collision-free designs when overfitting risk is inherently lower. Additionally, different domains' n-gram distribution differences may lead to different optimal strategies—code's n-gram semantics are more regular, collision noise costs may be higher; while natural language's ambiguity may make collision's ``soft clustering'' more valuable. Systematic cross-scale, cross-domain experiments will help clarify the applicability boundaries of collision effects.

\paragraph{Dynamic Training Strategies} This paper's revealed hot/cold flip phenomenon suggests: hot tier's advantage has an ``expiration date''—beneficial early, becoming a burden late. This suggests a natural improvement direction: \textbf{transforming static hot/cold strategy into dynamic training scheduling}. Specifically: (1) \textit{phased preference adjustment}—bias toward hot tier during warmup (letting the model quickly learn predictable patterns of high-frequency n-grams), gradually increasing cold tier weight or gate suppression ability in middle and late stages; (2) \textit{progressive regularization for hot embeddings}—allow hot embeddings to learn freely in early training, gradually introducing dropout or weight decay later to prevent overfitting; (3) \textit{changing hot tier from direct injection to indirect guidance}—for example, using hot embeddings as teacher signals or proposals, rather than directly adding to the residual stream. The core idea of these strategies is: acknowledging that ``collision-free'' accelerates overfitting, thus needing dynamic compensation during training.

\paragraph{Key Space Optimization and Feature Division of Labor} This paper focuses on the ``eliminating collisions'' optimization axis, but the more fundamental question may be: \textbf{which tokens should be merged, which should be distinguished, and what should different heads memorize}. The current tokenizer compression function $\mathcal{P}$ uses fixed rules (NFKC normalization, lowercasing, etc.), but this mapping could be data-driven and learnable—while maintaining hardenability and deployability. Furthermore, different attention heads may focus on different types of patterns (such as syntactic structures vs entity names vs phrase collocations), so exploring head-specific compression functions or n-gram selection strategies could let different heads' memory form complementary division of labor. This direction shifts research focus from ``more accurate indexing'' to ``more appropriate content,'' which may have higher ceiling than continuing to optimize $N_{\text{hot}}$ or table\_size.

\section{Conclusion}

This paper investigates a question that seems intuitively obvious but lacks empirical evidence: does eliminating collisions for high-frequency n-grams in Engram-style hashed memory improve training outcomes? At the scale we tested, the answer appears to be no. Under strict iso-parameter control, collision-free indexing did not yield significant improvements over collision-prone baselines.

Through route-stratified evaluation, we uncover two key findings. First, \textbf{collisions appear to act as implicit regularization}: collision-free configurations exhibit earlier overfitting on high-frequency positions, while collision-prone configurations delay this effect. Second, we identify a \textbf{gating credit assignment problem}: the gate learns early preferences that fail to adapt as training dynamics shift, creating a mismatch between gating weights and actual prediction performance.

The central implication is: \textbf{improving precision does not necessarily improve training benefits}. Before pursuing ``more precise'' retrieval, one should first verify whether precision is truly the bottleneck—otherwise, added complexity may destroy beneficial properties implicit in the original design. We hope the route-stratified evaluation methodology and the analytical framework proposed here will prove useful for diagnosing similar systems.

\bibliographystyle{unsrt}
\bibliography{refs}

\end{document}